\documentclass[10pt,twocolumn,letterpaper]{article}

\usepackage{cvpr}              

\usepackage{colortbl}
\usepackage{array}
\usepackage{amsmath}
\usepackage{multirow}
\usepackage{makecell}
\usepackage{graphicx}
\usepackage{algorithmicx,algorithm}
\usepackage[dvipsnames,svgnames]{xcolor}
\usepackage{svg}
\usepackage{alltt}

\usepackage{caption}
\usepackage{wrapfig}

\newcommand{\incre}[1]{\textcolor{teal!90}{#1}}

\definecolor{cvprblue}{rgb}{0.21,0.49,0.74}
\usepackage[pagebackref,breaklinks,colorlinks,citecolor=cvprblue]{hyperref}

\def\paperID{4667} 
\def\confName{CVPR}
\def\confYear{2024}

\title{PromptKD: Unsupervised Prompt Distillation for Vision-Language Models}

\author{
	Zheng Li\textsuperscript{\rm 1},
	Xiang Li\textsuperscript{\rm 2,1}\thanks{Corresponding author.},
	Xinyi Fu\textsuperscript{\rm 3},
	Xin Zhang\textsuperscript{\rm 1}, 
	Weiqiang Wang\textsuperscript{\rm 3},
        Shuo Chen\textsuperscript{\rm 4},
        Jian Yang\textsuperscript{\rm 1}\footnotemark[1]\\
	\textsuperscript{\rm 1} PCA Lab, VCIP, College of Computer Science, Nankai University \\ 
	\textsuperscript{\rm 2} NKIARI, Shenzhen Futian,
	\textsuperscript{\rm 3} Tiansuan Lab, Ant Group,
        \textsuperscript{\rm 4} RIKEN \\
	{\tt\small \{zhengli97, zhasion\}@mail.nankai.edu.cn, 
            \{xiang.li.implus, csjyang\}@nankai.edu.cn}\\
        {\tt\small \{fxy122992, weiqiang.wwq\}@antgroup.com, shuo.chen.ya@riken.jp}
}

\begin{document}
\maketitle

\begin{abstract}

Prompt learning has emerged as a valuable technique in enhancing vision-language models~(VLMs) such as CLIP for downstream tasks in specific domains. Existing work mainly focuses on designing various learning forms of prompts, neglecting the potential of prompts as effective distillers for learning from larger teacher models. In this paper, we introduce an unsupervised domain prompt distillation framework, which aims to transfer the knowledge of a larger teacher model to a lightweight target model through prompt-driven imitation using unlabeled domain images. Specifically, our framework consists of two distinct stages. In the initial stage, we pre-train a large CLIP teacher model using domain~(few-shot) labels. After pre-training, we leverage the unique decoupled-modality characteristics of CLIP by pre-computing and storing the text features as class vectors only once through the teacher text encoder.
In the subsequent stage, the stored class vectors are shared across teacher and student image encoders for calculating the predicted logits. Further, we align the logits of both the teacher and student models via KL divergence, encouraging the student image encoder to generate similar probability distributions to the teacher through the learnable prompts.
The proposed prompt distillation process eliminates the reliance on labeled data, enabling the algorithm to leverage a vast amount of unlabeled images within the domain.
Finally, the well-trained student image encoders and pre-stored text features~(class vectors) are utilized for inference. To our best knowledge, we are the first to (1) perform unsupervised domain-specific prompt-driven knowledge distillation for CLIP, and (2) establish a practical pre-storing mechanism of text features as shared class vectors between teacher and student. Extensive experiments on 11 datasets demonstrate the effectiveness of our method.
Code is publicly available at \url{https://github.com/zhengli97/PromptKD}.




%

\end{abstract}

\begin{figure}[t]
    \centering
    \includegraphics[width=0.82\linewidth]{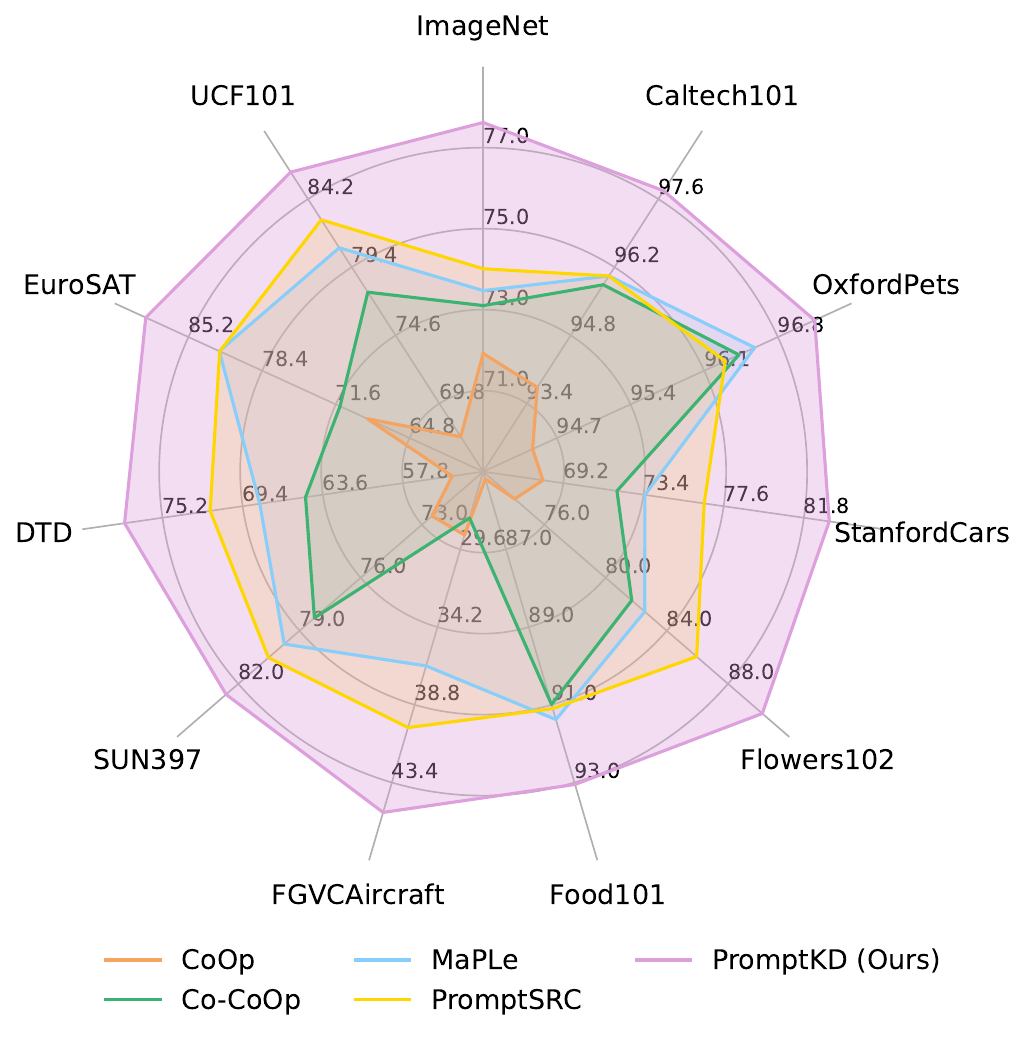}
    \vspace{-10pt}
    \caption{Harmonic mean~(HM) comparison on base-to-novel generalization. All methods adopt the \textbf{ViT-B/16 image encoder} from the pre-trained CLIP model. PromptKD achieves state-of-the-art performance on 11 diverse recognition datasets.}
    \label{fig:radar_compare}
    \vspace{-15pt}
\end{figure}

\section{Introduction}

Recently large pretrained vision-language models~(VLMs), such as CLIP~\cite{radford2021learning,zhang2023temo} and ALIGN~\cite{jia2021scaling}, have demonstrated superior generalization ability for domain-specific downstream tasks. Unlike conventional visual frameworks, the vision-language model, like CLIP, usually employs a two-tower architecture that includes an image encoder and a text encoder. These models are trained using a contrastive loss to learn a unified embedding space that aligns the representations of multi-modal signals.

\begin{figure*}[t]
    \flushright
    \includegraphics[width=0.86\linewidth]{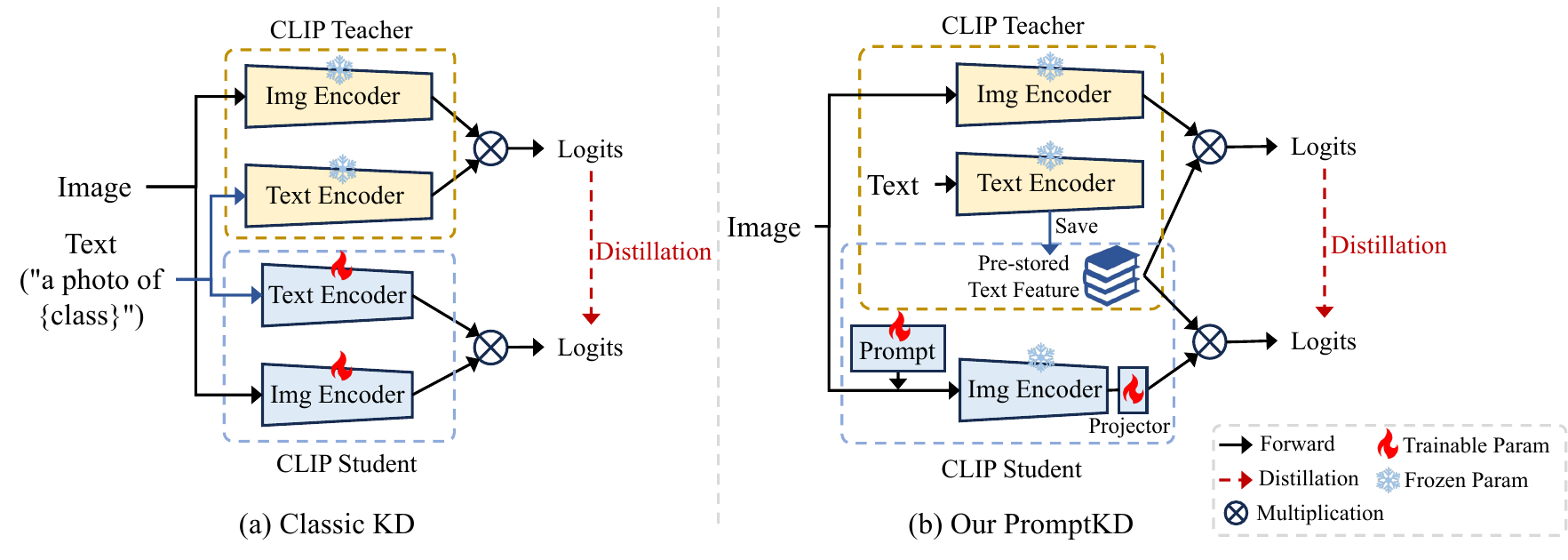}
    \vspace{-10pt}
    \caption{Architecture comparison between classic KD paradigm for CLIP (likewise CLIP-KD~\cite{yang2023clip}) and our prompt distillation framework. (a) Classic KD methods perform distillation between independent teacher and student models. Students are typically fully fine-tuned by teachers' soft labels. (b)~PromptKD breaks the rules of teacher-student independence. We propose to reuse the previously well-trained text features from the teacher pre-training stage and incorporate them into the student image encoder for both distillation and inference.
    }
    \label{fig:arch_compare}
    \vspace{-10pt}
\end{figure*}

To better optimize the models for domain-specific downstream tasks, various methods~\cite{zhou2022learning,zhou2022conditional,gao2023clip,khattak2023maple,yang2024fine} have been proposed to adapt the representation while keeping the original CLIP model fixed.
Inspired by the success of Nature Language Processing~(NLP)~\cite{lester2021power,li2021prefix} area, prompt learning~\cite{zhou2022learning,jia2022visual,zhou2022conditional} has been proposed to acquire continuous prompt representations as a replacement for meticulously designed hard prompts.
Based on the type of information learned by prompt, existing methods can be roughly divided into three types: text-based, visual-based, and both. Text-based methods~\cite{zhou2022learning,zhou2022conditional} propose to adaptively learn appropriate text prompts for downstream tasks, rather than fixed forms. Visual-based methods~\cite{jia2022visual,cheng2023e2vpt} follow similar principles and further apply them to visual modalities. Text-visual-based prompt methods~\cite{wang2022learning,khattak2023maple,lee2023read,khattak2023self} suggest a simultaneous learning strategy for prompts in both image and text branches, instead of treating them separately.


Prior research has primarily concentrated on acquiring effective \emph{formats of prompts} using scarce labeled data while preserving the outstanding generalization capabilities. In this paper, we introduce a novel unsupervised framework~(termed ``PromptKD'') where the prompt acts as a domain knowledge distiller, allowing the CLIP student model to absorb knowledge from a vast CLIP teacher model 
on extensive unlabeled domain data.
Specifically, our framework consists of two distinct stages: the teacher pre-training stage and the student distillation stage. 

In the initial stage, we first 
pre-train a large CLIP teacher model using existing advanced approaches~\cite{khattak2023maple,khattak2023self} on domain few-shot labeled data. After pre-training, we propose to leverage the unique decoupled-modality characteristics of CLIP by pre-computing and storing the text features as class vectors \emph{only once} through the teacher text encoder. 

In the subsequent stage, the stored class vectors are shared across the teacher and student image encoder to calculate the predicted logits without any extra computation costs from text branches. Different from the traditional knowledge distillation scheme where the weights of a student are usually fully tuned to mimic the teachers' statistical behavior as shown in Fig.~\ref{fig:arch_compare}(a), we propose to utilize the student's learnable visual prompts to align the logits of both teacher and student models via KL divergence, encouraging the student image encoder to generate similar probability distributions to the teacher through prompt distillation. 
Due to the dimensional differences between the features of teacher and student, an extra projector is implemented to adjust the features to account for the dimension disparity.

With the benefits of the teacher-student paradigm, we can leverage the pre-trained teacher to generate soft labels for unlabeled images from the target domain, thus enabling the training of students without the need for labeled images.
Finally, the well-trained student image encoder, along with the pre-stored teacher text features (class vectors), are employed for inference purposes. An architectural comparison of the classic distillation paradigm for CLIP and our proposed prompt distillation framework is illustrated in Fig.~\ref{fig:arch_compare}.

Experimental results in Fig.~\ref{fig:radar_compare} show that our PromptKD outperforms previous methods and achieves state-of-the-art performance on 11 diverse recognition datasets with the ViT-B/16 image encoder CLIP model. Specifically, our method achieves average improvements of 2.70\% and 4.63\% on the base and new classes on 11 diverse datasets.


Our contributions can be summarized as follows:
\begin{itemize}
    \item To our best knowledge, we are the first method to perform domain-specific prompt-based knowledge distillation for CLIP using unlabeled domain data.
    \item We leverage CLIP's unique decoupled-modality property to reuse pre-stored text features without incurring any additional computation costs from the text branch, thereby facilitating the distillation and inference processes.
    \item With the benefits of the teacher-student paradigm, we can utilize the teacher to generate soft labels on extensive unlabeled domain data, enabling the training of students without the need for labeled images.
    \item Extensive experiments on 11 datasets demonstrate the effectiveness of our method.
\end{itemize}

\section{Related Work}

\textbf{Prompt Learning in Vision-Language Models.}
Prompt learning is a technique that can transfer the large pre-trained model, like CLIP~\cite{radford2021learning}, towards downstream tasks~\cite{yao2022detclip,rao2022denseclip,ge2022domain} without the need for completely re-training the original model. It proposes to adapt the representations for specific tasks through learnable text or visual soft prompts instead of manually crafted hard prompts~(e.g., ``a photo of a \{classname\}"). Soft prompts~\cite{jia2022visual,zhou2022learning,zhou2022conditional,lee2023read,ren2023prompt} can be optimized by back-propagating through the frozen pre-trained model, resulting in better performance. 
Existing works mainly focus on designing various efficient forms of prompts using scarce labeled domain data. 
MaPLe~\cite{khattak2023maple} proposes to learn prompts for the image and text branches simultaneously, rather than a separate side. PromptSRC~\cite{khattak2023self} utilizes its original features to regularize the learning of prompts for each branch. 
Previous works necessitated forward and backward computations for each input in both image~\cite{dosovitskiy2020image,wang2022pvt} and text branches. In our work, we leverage the unique decoupled-modality characteristic of CLIP, saving well-trained teacher text features as class vectors for student distillation. In this way, the training of student CLIP is simplified to solely include forward and backward calculations of the image branch, without requiring the text branch. 


\noindent\textbf{Zero-shot Learning.}
Given the labeled training set of the seen classes, zero-shot learning~(ZSL)~\cite{xian2018zero,wang2019survey,liu2023meaningful} aims to learn a classifier that can classify testing samples of unseen classes.
Existing methods can be roughly divided into two types based on whether test images are available: Inductive~\cite{zhang2017learning,xian2018feature} and Transductive~\cite{song2018transductive,wan2019transductive} ZSL. Previous works on prompt learning, such as MaPLe and PromptSRC, have mainly focused on the instance inductive settings where only labeled training instances are available.
In our paper, we explore the transductive ZSL setting where both seen and unseen class images are all utilized in model learning. Specifically, our teacher model follows the same training scheme as PromptSRC, which is trained on samples from seen classes with ground truth labels. The difference is that the target student model is trained on the full unlabeled dataset, which contains all samples of both seen and unseen classes, without using any ground truth labels. 

\noindent\textbf{Knowledge Distillation.}
Knowledge distillation~\cite{hinton2015distilling} aims to train a lightweight student model under the supervision of a large pretrained teacher model. In recent years, various distillation forms have emerged for effective knowledge transfer from teachers to students, such as logits alignment~\cite{zhang2018deep,li2020online,zhao2022decoupled,li2023curriculum}, feature imitation~\cite{yang2021knowledge,chen2022knowledge,li2023mask} and sample relationship matching~\cite{park2019relational,yang2022mutual}. 
In addition to traditional image classification topics, knowledge distillation has achieved great success in many vision tasks, including object detection~\cite{cao2022pkd,wang2023crosskd,jia2024mssd}, image segmentation~\cite{liu2019structured,yang2022cross}, and pose estimation~\cite{li2021online}. Recently, many works~\cite{pei2023clipping,yang2023clip,wu2023tinyclip,laroudie2023improving} have turned their attention to the CLIP model. These works propose leveraging the CLIP model's exceptional generalization capabilities to enhance the learning of existing models. 
CLIP-KD~\cite{yang2023clip} find that in distilling pre-trained CLIP models, the simplest feature mimicry with the MSE loss approach yields the best results. TinyCLIP~\cite{wu2023tinyclip} performs cross-modal feature alignment in affinity space between teacher and student. 
Our approach differs from previous distillation methods that train the \emph{entire student} model by leveraging a pre-trained large CLIP teacher. In our work, we employ a more efficient approach by utilizing \emph{student prompts} for distillation while keeping the student's original CLIP weights frozen. This allows us to achieve the desired knowledge transfer without the need for extensive re-training of the student model.


\begin{figure*}[t]
    \centering
    \includegraphics[width=0.99\linewidth]{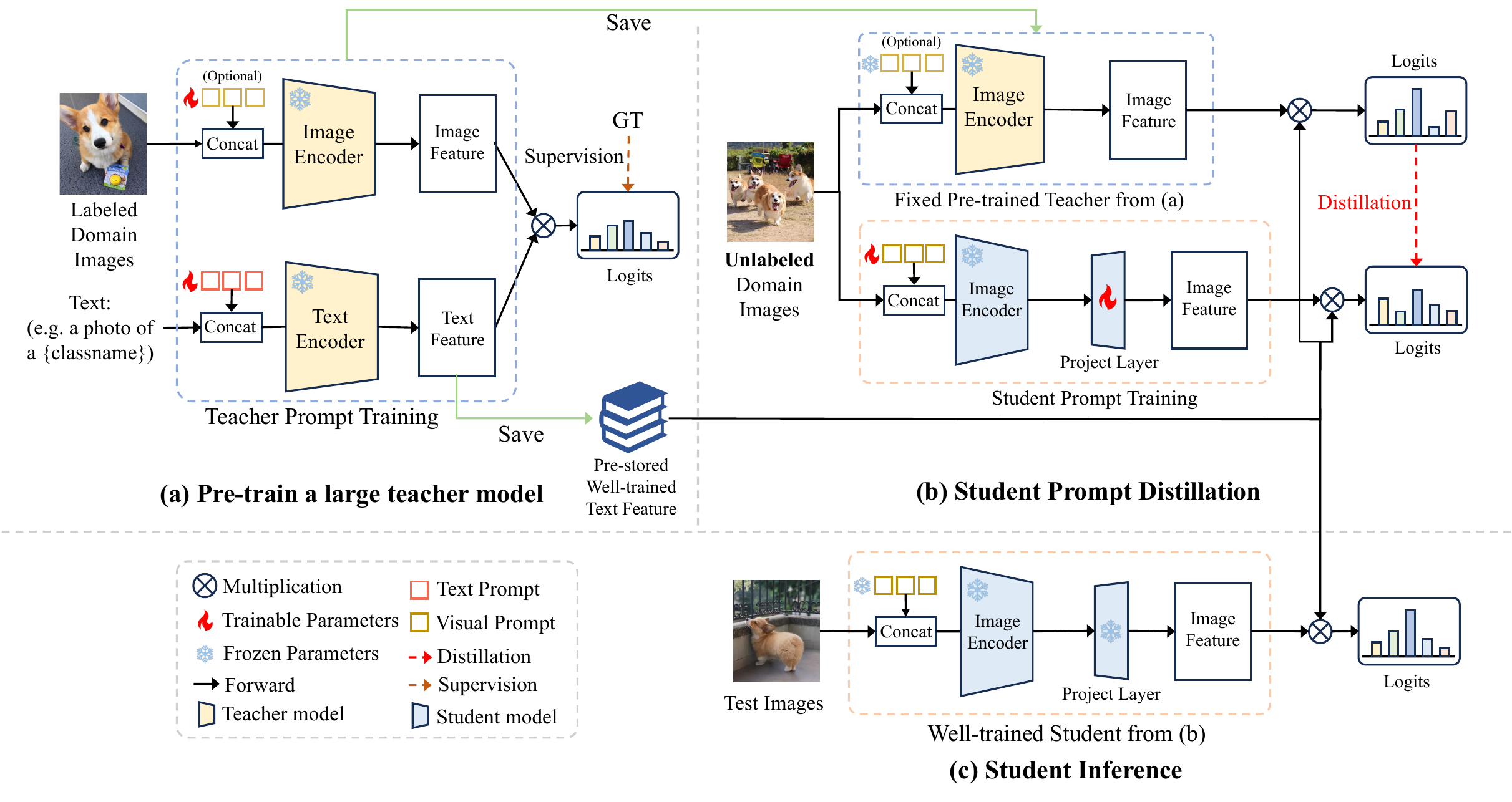}
    \vspace{-10pt}
    \caption{An overview of our proposed prompt distillation~(PromptKD) framework. (a) We first pre-train a large CLIP teacher model using existing state-of-the-art prompt learning methods with labeled training images. Then we save the well-trained text features of all possible classes for the next stages. (b) During the distillation stage, the training is focused on student image prompts and the project layer, and there are no extra computational expenses associated with the text encoding process when utilizing the pre-saved text features as class vectors. (c) Finally, the well-trained student and pre-stored class vectors are utilized for inference. 
    }  
    \label{fig:framework}
    \vspace{-10pt}
\end{figure*}

\section{Method}

Prompt learning~\cite{zhou2022learning,jia2022visual} aims to enhance the performance of existing VLMs like CLIP to downstream tasks by incorporating learnable prompts. Existing works mainly focus on devising effective learning formats of prompts using scarce labeled domain data while ensuring strong generalization capabilities to unseen images.
In this paper, we first explore prompts as an effective knowledge distiller, allowing the CLIP student model to learn from the large CLIP teacher model by aligning their predictions on extensive unlabeled domain images. An overview of our proposed prompt distillation method is illustrated in Fig.~\ref{fig:framework}. Specifically, our method comprises two main stages: the teacher pre-training stage and the student prompt distillation stage. In the initial stage, we first pre-train a large CLIP teacher model using existing advanced approaches on few-shot labeled data, as depicted in Fig.~\ref{fig:framework}(a). After pre-training, we extract and preserve the highly proficient text features obtained from the teacher text encoder as class vectors. In the subsequent stage, the pre-stored class vectors are effectively reused by multiplying them with the outputs of both the teacher and student image encoders, resulting in predictions for each model. Then we initiate the distillation process by promoting prompt imitation, encouraging the student model to generate similar predictions to the teacher model, as illustrated in Fig.~\ref{fig:framework}(b). An additional projector is introduced to align the dimensions of teacher text features and student image features. 
Finally, the well-trained student image encoder branch and pre-stored teacher text features~(class vectors) are utilized for inference~(see Fig.~\ref{fig:framework}(c)).

Below we first introduce the background knowledge of VLMs and the knowledge distillation method in Sec.~\ref{section:background}. Then we introduce our method in detail in Sec.~\ref{section:promptkd}.

\subsection{Background}
\label{section:background}
\noindent\textbf{Vision-Language Models.} 
Existing VLMs like CLIP~\cite{radford2021learning} and ALIGN~\cite{jia2021scaling} are designed to align images and texts in order to learn a joint embedding space. Following~\cite{zhou2022conditional,khattak2023maple,khattak2023self}, we consider CLIP as our foundation model. Specifically, CLIP consists of two encoders, one for image and the other for text. Given a labeled visual recognition dataset $D=\{x_{j}, y_{j}\}_{j=1}^{M}$ that includes a set of $N$ class names $\text{c}=\{c_{i}\}_{i=1}^{N}$, CLIP generates textual descriptions $t_{i}$ using the template ``a photo of a \{$c_{i}$\}" for each class name. Then each text description ${t_{i}}$ is fed into the text encoder ${f_{T}}$ to obtain the normalized text feature $w_{i}=f_{T}(t_{i})/||f_{T}(t_{i})||_{2}\in \mathbb{R}^{d}$, where $d$ represents the feature dimension. 
The complete text features $\text{W}=[w_{1}, w_{2}, ..., w_{N}]\in \mathbb{R}^{N\times d}$ of all classes can be considered as the classification weight vector for classifying an image. Given an input image $x$ from the dataset $D$, the image encoder $f_{I}$ takes as input and generates the normalized image feature $u=f_{I}(x)/||f_{I}(x)||_{2}\in\mathbb{R}^{d}$. 
The output probability is calculated as follows:
\begin{equation}
    p(y|x) = \frac{\text{exp}(u w_{y}^{\mathsf{T}}/\tau)}{\sum_{i=1}^{N}\text{exp}(u w_{i}^{\mathsf{T}}/\tau)},
\label{equation:output_prob}
\end{equation}
where $uw^{\mathsf{T}}$ represent the output logit and $\tau$ is the temperature parameter.

Instead of manually crafted hard prompts, recent works like CoOp~\cite{zhou2022learning} propose to adaptively learn appropriate soft \emph{textual} prompts for downstream tasks. Concretely, $M$ learnable textual vectors $\{v_{1}, v_{2}, ..., v_{M}\}$, i.e., prefix, are added before the CLASS token to create a contextualized representation. Then the prompt $t_{i}$ for class $c_{i}$ becomes $t_{i}=\{v_{1}, v_{2}, ..., v_{M}, c_{i}\}$, where each vector $v_{i}~(i\in1, 2, ..., M)$ have the same dimension with the word embeddings and $M$ is a hyperparameter that determines the length of the prefix.
In addition to text prompt tuning methods, visual prompts have also been extensively explored. Some works~\cite{jia2022visual,khattak2023maple,khattak2023self} follow the same idea as the text prompt method, adding multiple learnable visual prefixes to the image patch as input to the image encoder. These visual prompts aim to guide the image encoder to extract more meaningful and task-relevant visual features. By incorporating these learnable visual prefixes, the model can leverage additional context and prior knowledge to improve its performance on image understanding tasks.

\noindent\textbf{Knowledge Distillation.}
Originally proposed by Hinton et al.~\cite{hinton2015distilling}, knowledge distillation aims to transfer the knowledge of a pretrained heavy teacher model to a lightweight student model. 
After the distillation, the student can master the expertise of the teacher and be used for final deployment. Specifically, the Kullback-Leibler~(KL) divergence loss is utilized to match the output distribution of two models, which can be formulated as follows:
\begin{equation}
L_{kd}(q^{t}, q^{s}, \tau) = \tau^{2} KL(\sigma (q^{t}/\tau),\sigma (q^{s}/\tau)).
\label{equation:kd}
\end{equation}
where $q^{t}$ and $q^{s}$ denote the logits predicted by the teacher and student. $\sigma(\cdot)$ is the softmax function and $\tau$ is the temperature~\cite{hinton2015distilling,li2023curriculum} which controls the softness of distribution. 

\subsection{PromptKD: Prompt Distillation for VLMs}
\label{section:promptkd}
Our proposed prompt distillation framework comprises two stages: teacher pre-training and student prompt distillation, as illustrated in Fig.~\ref{fig:framework}. In this section, we provide a comprehensive explanation of each stage.

\noindent\textbf{Stage I: Teacher Pretraining.} In the initial stage, we begin by pre-training a large CLIP teacher model using labeled domain data, as illustrated in Fig.~\ref{fig:framework}(a). To accomplish this, we can employ existing prompt learning methods such as MaPLe~\cite{khattak2023maple} and PromptSRC~\cite{khattak2023self}, or alternatively, utilize a publicly available pretrained CLIP model for simplicity. Given a labeled domain dataset $D_{labeled}=\{x_{i}, y_{i}\}_{i=1}^{M}$ with a set class name, the teacher CLIP model takes training images and text descriptions with category names as input, and passes through the image encoder $f_{I}^{t}$ and text encoder $f_{T}^{t}$ to obtain the corresponding normalized image features $u\in \mathbb{R}^{d}$ and text features $w\in \mathbb{R}^{d}$. The final output result $p^{t}$ is calculated by Eqn.~\eqref{equation:output_prob}. Typically, the parameters of teacher soft prompts are updated by minimizing the cross-entropy loss between predicted probabilities $p$ and ground truth labels $y$.

Once the training of the text encoder is completed, the output features remain fixed and do not require further updates. In this case, we save the well-trained teacher text features of all $N$ classes $W=[w_{1}, w_{2}, ..., w_{N}]\in \mathbb{R}^{N\times d}$ as \emph{shared} class vectors that will be utilized in the subsequent stages of the process. This operation eliminates the necessity of having the student CLIP text branch, resulting in substantial computational cost savings during the training process. In addition, through our PromptKD method, we can replace the large teacher's heavy image encoder with a student lightweight image encoder, reducing the computational cost during deployment while maintaining competitive performance.



\noindent\textbf{Stage II: Student Prompt Distillation.}
At this stage, we aim to train a student model by encouraging the student to align with the teacher's output results through prompt imitation, as shown in Fig.~\ref{fig:framework}(b). 
Thanks to the strategy of reusing teacher text features, 
we only need to train the student image encoder branch $f_{I}^{s}$ with learnable visual prompts and the feature projector. 
In the context of an unlabeled domain dataset $D_{unlabeled}$, by inputting the image $x$ into both the pre-trained teacher's and the untrained student's image branches, we can acquire the normalized teacher image features $u^{t}=f_{I}^{t}(x)/||f_{I}^{t}(x)||_{2}\in \mathbb{R}^{d}$ and student image features $u^{s}=P(f^{s}_{I}(x))/||P(f^{s}_{I}(x))||_{2}\in \mathbb{R}^{d}$. The learnable projector $P(\cdot)$ in the student image encoder branch is introduced to match the feature dimensions at a relatively small cost while being effective enough to ensure accurate alignment. Then we multiply the pre-stored teacher text features $W\in \mathbb{R}^{N\times d}$ with the teacher and student image features to obtain the output logits $q^{t}=u^{t}W^{\mathsf{T}}\in \mathbb{R}^{N}$ and $q^{s}=u^{s}W^{\mathsf{T}}\in \mathbb{R}^{N}$, respectively. 
We optimize the student model to produce similar output to the teacher model on the unlabeled domain dataset $D_{unlabeled}$, which can be formulated as follows:
\begin{equation}
    L_{stu}=L_{kd}(q^{t}, q^{s}, \tau).
\end{equation}

Algorithm~\ref{algo:promoptkd_algo} provides PromptKD’s PyTorch-style pseudocode.

\noindent\textbf{Inference.} Finally, the well-trained student image encoder $f_{I}^{s}$, along with the pre-stored teacher text features $W$ (class vectors), are employed for inference purposes.



    


\begin{algorithm}[t]
    \caption{Pseudocode of PromptKD in PyTorch.}
    \label{algo:promoptkd_algo}
    \footnotesize
    \vspace{-10pt}
    \begin{alltt}\color{ForestGreen}
# tea\_t: text encoder of teacher CLIP
# tea\_i: image encoder of teacher CLIP
# stu\_i: image encoder of student CLIP
# l\_tea: teacher output logits
# l\_stu: student output logits
# Proj: Feature Projector
 
\color{ForestGreen}# init \color{Black}
f_txt_t = tea_t(txt_of_all_classes)

\color{ForestGreen}# forward \color{Black}
for img in unlabeled_dataset:
    f_img\_t = tea_i(img)
    f_img\_s = stu_i(img)
    
    f_img_s = Proj(f_img_s)

    \color{ForestGreen}# get output predictions \color{Black}
    l\_tea = f\_img\_t * f\_txt\_t.t()
    l\_stu = f\_img\_s * f\_txt\_t.t()
    
    \color{ForestGreen}# calculate distillation loss \color{Black}
    loss = KLDivergence(l\_stu, l\_tea)
    loss.backward()
\end{alltt}
\vspace{-10pt}
\end{algorithm}

\begin{table*}[t]
    \centering
    \begin{subtable}[t]{0.32\linewidth}
        \centering
        \resizebox{0.91\linewidth}{!}
        {
        \begin{tabular}{cccc}
            \hline\noalign{\smallskip}
            ViT-B/16 & Base  & Novel & HM \\
            \hline\noalign{\smallskip}
            CLIP    & 69.34 & 74.22 & 71.70 \\
            CoOp    & 82.69 & 63.22 & 71.66 \\
            CoCoOp  & 80.47 & 71.69 & 75.83 \\
            MaPLe   & 82.28 & 75.14 & 78.55 \\
            PromptSRC & 84.26 & 76.10 & 79.97 \\
            \hline\noalign{\smallskip}
            \cellcolor{lightgray!30}PromptKD & \cellcolor{lightgray!30}86.96 & \cellcolor{lightgray!30}80.73 & \cellcolor{lightgray!30}83.73 \\
            $\Delta$ & \incre{+2.70} & \incre{+4.63} & \incre{+3.76} \\
            \hline
        \end{tabular}
        }
        \caption{Average over 11 datasets.}
    \end{subtable}
    \begin{subtable}[t]{0.32\linewidth}
        \centering
        \resizebox{0.91\linewidth}{!}
        {
        \begin{tabular}{cccc}
            \hline\noalign{\smallskip}
            ViT-B/16 & Base  & Novel & HM \\
            \hline\noalign{\smallskip}
            CLIP & 72.43 & 68.14 & 70.22 \\
            CoOp & 76.47 & 67.88 & 71.92 \\
            CoCoOp & 75.98 & 70.43 & 73.10 \\
            MaPLe  & 76.66 & 70.54 & 73.47 \\
            PromptSRC & 77.60 & 70.73 & 74.01 \\
            \hline\noalign{\smallskip}
            \cellcolor{lightgray!30}PromptKD & \cellcolor{lightgray!30}80.83 & \cellcolor{lightgray!30}74.66 & \cellcolor{lightgray!30}77.62 \\
            $\Delta$ & \incre{+3.23} & \incre{+3.93} & \incre{+3.61} \\
            \hline
        \end{tabular}
        }
        \caption{ImageNet}
    \end{subtable}
    \begin{subtable}[t]{0.32\linewidth}
        \centering
        \resizebox{0.91\linewidth}{!}
        {
        \begin{tabular}{cccc}
            \hline\noalign{\smallskip}
            ViT-B/16 & Base  & Novel & HM \\
            \hline\noalign{\smallskip}
            CLIP & 96.84 & 94.00 & 95.40 \\
            CoOp & 98.00 & 89.81 & 93.73 \\
            CoCoOp & 97.96 & 93.81 & 95.84 \\
            MaPLe  & 97.74 & 94.36 & 96.02 \\
            PromptSRC & 98.10 & 94.03 & 96.02 \\
            \hline\noalign{\smallskip}
            \cellcolor{lightgray!30} PromptKD & \cellcolor{lightgray!30}98.91 & \cellcolor{lightgray!30}96.65 & \cellcolor{lightgray!30}97.77 \\
            $\Delta$ & \incre{+0.81} & \incre{+2.62} & \incre{+1.75} \\
            \hline
        \end{tabular}
        }
        \caption{Caltech101}
    \end{subtable}  
    
    \vspace{5pt}
    
    \begin{subtable}[t]{0.32\linewidth}
        \centering
        \resizebox{0.91\linewidth}{!}
        {
        \begin{tabular}{cccc}
            \hline\noalign{\smallskip}
            ViT-B/16 & Base  & Novel & HM \\
            \hline\noalign{\smallskip}
            CLIP & 91.17 & 97.26 & 94.12 \\
            CoOp & 93.67 & 95.29 & 94.47 \\
            CoCoOp & 95.20 & 97.69 & 96.43 \\
            MaPLe  & 95.43 & 97.76 & 96.58 \\
            PromptSRC & 95.33 & 97.30 & 96.30 \\
            \hline\noalign{\smallskip}
            \cellcolor{lightgray!30}PromptKD & \cellcolor{lightgray!30}96.30 & \cellcolor{lightgray!30}98.01 & \cellcolor{lightgray!30}97.15 \\
            $\Delta$ & \incre{+0.97} & \incre{+0.71} & \incre{+0.85} \\
            \hline
        \end{tabular}
        }
        \caption{OxfordPets}  
    \end{subtable}
    \begin{subtable}[t]{0.32\linewidth}
        \centering
        \resizebox{0.91\linewidth}{!}
        {
        \begin{tabular}{cccc}
            \hline\noalign{\smallskip}
            ViT-B/16 & Base  & Novel & HM \\
            \hline\noalign{\smallskip}
            CLIP & 63.37 & 74.89 & 68.65 \\
            CoOp & 78.12 & 60.40 & 68.13 \\
            CoCoOp & 70.49 & 73.59 & 72.01 \\
            MaPLe & 72.94 & 74.00 & 73.47 \\
            PromptSRC & 78.27 & 74.97 & 76.58 \\
            \hline\noalign{\smallskip}
            \cellcolor{lightgray!30}PromptKD & \cellcolor{lightgray!30}82.80 & \cellcolor{lightgray!30}83.37 & \cellcolor{lightgray!30}83.13 \\
            $\Delta$ & \incre{+4.53} & \incre{+8.40} & \incre{+6.55} \\
            \hline
        \end{tabular}
        }
        \caption{StanfordCars}  
    \end{subtable}
    \begin{subtable}[t]{0.32\linewidth}
        \centering
        \resizebox{0.91\linewidth}{!}
        {
        \begin{tabular}{cccc}
            \hline\noalign{\smallskip}
            ViT-B/16 & Base  & Novel & HM \\
            \hline\noalign{\smallskip}
            CLIP & 72.08 & 77.80 & 74.83 \\
            CoOp & 97.60 & 59.67 & 74.06 \\
            CoCoOp & 94.87 & 71.75 & 81.71 \\
            MaPLe & 95.92 & 72.46 & 82.56 \\
            PromptSRC & 98.07 & 76.50 & 85.95 \\
            \hline\noalign{\smallskip}
            \cellcolor{lightgray!30}PromptKD & \cellcolor{lightgray!30}99.42 & \cellcolor{lightgray!30}82.62 & \cellcolor{lightgray!30}90.24 \\
            $\Delta$ & \incre{+1.35} & \incre{+6.12} & \incre{+4.29} \\
            \hline
        \end{tabular}
        }
        \caption{Flowers102} 
    \end{subtable}

    \vspace{5pt}

    \begin{subtable}[t]{0.32\linewidth}
        \centering
        \resizebox{0.91\linewidth}{!}
        {
        \begin{tabular}{cccc}
            \hline\noalign{\smallskip}
            ViT-B/16 & Base  & Novel & HM \\
            \hline\noalign{\smallskip}
            CLIP & 90.10 & 91.22 & 90.66 \\
            CoOp & 88.33 & 82.26 & 85.19 \\
            CoCoOp & 90.70 & 91.29 & 90.99 \\
            MaPLe & 90.71 & 92.05 & 91.38 \\
            PromptSRC & 90.67 & 91.53 & 91.10 \\
            \hline\noalign{\smallskip}
            \cellcolor{lightgray!30}PromptKD & \cellcolor{lightgray!30}92.43 & \cellcolor{lightgray!30}93.68 & \cellcolor{lightgray!30}93.05 \\
            $\Delta$ & \incre{+1.76} & \incre{+2.15} & \incre{+1.95} \\
            \hline
        \end{tabular}
        }
        \caption{Food101} 
    \end{subtable}
    \begin{subtable}[t]{0.32\linewidth}
        \centering
        \resizebox{0.91\linewidth}{!}
        {
        \begin{tabular}{cccc}
            \hline\noalign{\smallskip}
            ViT-B/16 & Base  & Novel & HM \\
            \hline\noalign{\smallskip}
            CLIP & 27.19 & 36.29 & 31.09 \\
            CoOp & 40.44 & 22.30 & 28.75 \\
            CoCoOp & 33.41 & 23.71 & 27.74 \\
            MaPLe & 37.44 & 35.61 & 36.50 \\
            PromptSRC & 42.73 & 37.87 & 40.15 \\
            \hline\noalign{\smallskip}
            \cellcolor{lightgray!30}PromptKD & \cellcolor{lightgray!30}49.12 & \cellcolor{lightgray!30}41.81 & \cellcolor{lightgray!30}45.17  \\
            $\Delta$ & \incre{+6.39} & \incre{+3.94} & \incre{+5.02} \\
            \hline
        \end{tabular}
        }
        \caption{FGVCAircraft} 
    \end{subtable}
    \begin{subtable}[t]{0.32\linewidth}
        \centering
        \resizebox{0.91\linewidth}{!}
        {
        \begin{tabular}{cccc}
            \hline\noalign{\smallskip}
            ViT-B/16 & Base  & Novel & HM \\
            \hline\noalign{\smallskip}
            CLIP & 69.36 & 75.35 & 72.23 \\
            CoOp & 80.60 & 65.89 & 72.51 \\
            CoCoOp & 79.74 & 76.86 & 78.27 \\
            MaPLe & 80.82 & 78.70 & 79.75 \\
            PromptSRC & 82.67 & 78.47 & 80.52 \\
            \hline\noalign{\smallskip}
            \cellcolor{lightgray!30}PromptKD & \cellcolor{lightgray!30}83.69 & \cellcolor{lightgray!30}81.54 & \cellcolor{lightgray!30}82.60 \\
            $\Delta$ & \incre{+1.02} & \incre{+3.07} & \incre{+2.08} \\
            \hline
        \end{tabular}
        }
        \caption{SUN397}  
    \end{subtable}

    \vspace{5pt}

    \begin{subtable}[t]{0.32\linewidth}
        \centering
        \resizebox{0.91\linewidth}{!}
        {
        \begin{tabular}{cccc}
            \hline\noalign{\smallskip}
            ViT-B/16 & Base  & Novel & HM \\
            \hline\noalign{\smallskip}
            CLIP & 53.24 & 59.90 & 56.37 \\
            CoOp & 79.44 & 41.18 & 54.24 \\ 
            CoCoOp & 77.01 & 56.00 & 64.85 \\
            MaPLe & 80.36 & 59.18 & 68.16 \\
            PromptSRC & 83.37 & 62.97 & 71.75 \\
            \hline\noalign{\smallskip}
            \cellcolor{lightgray!30}PromptKD & \cellcolor{lightgray!30}85.84 & \cellcolor{lightgray!30}71.37 & \cellcolor{lightgray!30}77.94 \\
            $\Delta$ & \incre{+2.47} & \incre{+8.40} & \incre{+6.19} \\
            \hline
        \end{tabular}
        }
        \caption{DTD}  
    \end{subtable}
    \begin{subtable}[t]{0.32\linewidth}
        \centering
        \resizebox{0.91\linewidth}{!}
        {
        \begin{tabular}{cccc}
            \hline\noalign{\smallskip}
            ViT-B/16 & Base  & Novel & HM \\
            \hline\noalign{\smallskip}
            CLIP & 56.48 & 64.05 & 60.03 \\
            CoOp & 92.19 & 54.74 & 68.69 \\
            CoCoOp & 87.49 & 60.04 & 71.21 \\
            MaPLe & 94.07 & 73.23 & 82.35 \\
            PromptSRC & 92.90 & 73.90 & 82.32 \\
            \hline\noalign{\smallskip}
            \cellcolor{lightgray!30}PromptKD & \cellcolor{lightgray!30}97.54 & \cellcolor{lightgray!30}82.08 & \cellcolor{lightgray!30}89.14\\
            $\Delta$ & \incre{+4.64} & \incre{+8.18} & \incre{+6.82} \\
            \hline
        \end{tabular}
        }
        \caption{EuroSAT} 
    \end{subtable}
    \begin{subtable}[t]{0.32\linewidth}
        \centering
        \resizebox{0.91\linewidth}{!}
        {
        \begin{tabular}{cccc}
            \hline\noalign{\smallskip}
            ViT-B/16 & Base  & Novel & HM \\
            \hline\noalign{\smallskip}
            CLIP & 70.53 & 77.50 & 73.85 \\
            CoOp & 84.69 & 56.05 & 67.46 \\
            CoCoOp & 82.33 & 73.45 & 77.64 \\
            MaPLe & 83.00 & 78.66 & 80.77 \\
            PromptSRC & 87.10 & 78.80 & 82.74 \\
            \hline\noalign{\smallskip}
            \cellcolor{lightgray!30}PromptKD & \cellcolor{lightgray!30}89.71 & \cellcolor{lightgray!30}82.27 & \cellcolor{lightgray!30}86.10 \\
            $\Delta$ & \incre{+2.61} & \incre{+3.47} & \incre{+3.36} \\
            \hline
        \end{tabular}
        }
        \caption{UCF101}
    \end{subtable}
    \vspace{-5pt}
    \caption{Comparison with existing state-of-the-art methods on base-to-novel generalization. Our proposed PromptKD demonstrates strong generalization ability and achieves significant improvements on 11 recognition datasets given the \textbf{ViT-B/16 image encoder} of the CLIP model. In our approach, the default teacher model is the ViT-L/14 CLIP model. The symbol $\Delta$ denotes the performance improvement compared to the previous SOTA method PromptSRC. Our PromptKD outperforms previous methods on all datasets.}
    \label{table:base2novel}
    \vspace{-7pt}
\end{table*}

\begin{table*}[t]
    \centering
    \resizebox{0.97\linewidth}{!}
    {
        \begin{tabular}{ccccccccccccc}
        \hline\noalign{\smallskip}
        ~   & ~         & \multicolumn{10}{c}{Target Dataset} \\
        \noalign{\smallskip}
        \multirow{2}*{ZSL} & \multirow{2}*{ViT-B/16} & Caltech & Oxford & Standford & Flowers & \multirow{2}*{Food101} & FGVC & \multirow{2}*{SUN397} & \multirow{2}*{DTD} & Euro & \multirow{2}*{UCF101} & \multirow{2}*{Avg.} \\
        ~   &          & 101     & Pets   & Cars      & 102     & ~  & Aircraft & ~ & ~     & SAT  & ~ & ~    \\
        \hline\noalign{\smallskip}
        ~ & CoOp       & 93.70 & 89.14 & 64.51 & 68.71 & 85.30 & 18.47 & 64.15 & 41.92 & 46.39 & 66.55 & 63.88 \\
        In- & CoCoOp  & \textbf{94.43} & 90.14 & 65.32 & 71.88 & 86.06 & 22.94 & 67.36 & 45.73 & 45.37 & 68.21 & 65.74 \\
        ductive & MaPLe & 93.53 & 90.49 & 65.57 & 72.23 & 86.20 & 24.74 & 67.01 & 46.49 & 48.06 & 68.69 & 66.30 \\
        ~ & PromptSRC   & 93.60 & 90.25 & 65.70 & 70.25 & 86.15 & 23.90 & 67.10 & 46.87 & 45.50 & 68.75 & 65.81 \\
        \hline\noalign{\smallskip}
        Trans- & PromptKD  & 93.61 & \textbf{91.59} & \textbf{73.93} & \textbf{75.33} & \textbf{88.84} & \textbf{26.24} & \textbf{68.57} & \textbf{55.08} & \textbf{63.74} & \textbf{76.39} & \textbf{71.33} \\
        ductive & $\Delta$ & \incre{+0.01} & \incre{+1.34} & \incre{+8.23} & \incre{+5.08} & \incre{+2.69} & \incre{+2.34} & \incre{+1.47} & \incre{+8.21} & \incre{+18.24} & \incre{+7.64} & \incre{+5.52} \\
        \hline
        \end{tabular}
    }
    \caption{Comparison of PromptKD with existing advanced approaches on cross-dataset benchmark evaluation. Based on our pipeline, we perform unsupervised prompt distillation using the unlabeled domain data respectively (i.e., the transductive setting). The source model is trained on ImageNet~\cite{deng2009imagenet}. ``ZSL'' denotes the setting type for Zero-Shot Learning. PromptKD achieves better results on 9 of 10 datasets.
    }
    \label{table:cross_dataset}
    \vspace{-10pt}
\end{table*}

\section{Experiments}

\subsection{Settings}


\noindent\textbf{Base-to-novel Generalization.} 
Following~\cite{zhou2022conditional,khattak2023self,khattak2023maple}, we split the training and testing datasets into base and novel classes. 
The teacher is pre-trained using the PrompSRC~\cite{khattak2023self} method, following the same training setting as PromptSRC.
During distillation, we use the entire unlabeled training set to train our students.
After distillation, the student's performance on the base and the novel class is evaluated on the testing set. 

\noindent\textbf{Cross-dataset Evaluation.}
Same as PromptSRC~\cite{khattak2023self}, our teacher model is pre-trained on the source dataset~(i.e., ImageNet) with a 16-shot training data configuration. 
Then we use the training set of unlabeled target datasets to train students and evaluate their performance on the test set after training.
In PromptKD, we use unlabeled images of unseen classes for student training which belongs to the \emph{transductive} zero-shot learning method. For previous methods such as CoOp, MaPLe, and PromptSRC, their training is based on seen class data and belongs to the \emph{inductive} paradigm. 

\noindent\textbf{Datasets.} We evaluate the model performance on 11 popular recognition datasets. The details of each dataset are attached in the Appendix.


\noindent\textbf{Implementation Details.}
We use the ViT-L/14 CLIP model as our teacher model and the ViT-B/16 CLIP model as our target student model. Unless otherwise stated, the PromptSRC~\cite{khattak2023self} is leveraged as our default method to pre-train our teacher model. 
We report base and novel class accuracy and their harmonic mean~(HM) averaged over 3 runs.
Due to page limitations, please refer to the Appendix for more implementation details and experimental results.

\subsection{Base-to-novel Generalization}

As shown in Table~\ref{table:base2novel}, based on the same ViT-B/16 image encoder of the pre-trained CLIP, we compare the performance of our proposed PromptKD with recent state-of-the-art prompt learning
methods including CoOp, CoCoOp, MaPLe and PromptSRC on 11 recognition datasets. 
In comparison with previous works, PromptKD shows superior performance on all 11 datasets. 
The accuracy of our pre-trained teacher model with ViT-L/14 image encoder on each dataset is provided in the Appendix.

\subsection{Cross-dataset Evaluation}

Table~\ref{table:cross_dataset} shows the performance comparison between CoOp, CoCoOp, MaPLe, PromoptSRC, and PromptKD. In comparison with previous methods, our method demonstrates better performance on 9 of 10 datasets.
leading to an average improvement of 5.52\% over the previous method. 

\subsection{Comparison with Other Methods}


In PromptKD, we utilize unlabeled images to train the target student model. Table~\ref{table:method_comparison} presents a comprehensive comparison between our method and other recent methods that also leverage unlabeled data for model training. While many methods resort to pseudo-labeling of unlabeled data for training, our approach adopts a teacher-student paradigm. In this paradigm, the teacher model plays a pivotal role by furnishing soft labels to train the student models on the unlabeled data. For fair comparisons, the methods using few-shot labels (\cite{menghini2023enhancing} and PromptKD) are all implemented based on PromptSRC framework. All experiments utilize \textbf{ViT-B/16} CLIP with the few-shot number being 16. The results on Flowers102 underscore the clear performance advantages of our approach over previous methods.

\begin{table}[t]
    \centering
    \resizebox{0.87\linewidth}{!}
    {
        \begin{tabular}{ccccc}
        \hline\noalign{\smallskip}
        Method    & Domain Data & Base  & Novel & HM  \\
        \hline\noalign{\smallskip}
        CLIP      & Zero-shot & 72.08 & 77.80 & 74.83 \\
        PromptSRC & Few-shot  & 98.07 & 76.50 & 85.95 \\
        \hline\noalign{\smallskip}
        CLIP-PR~\cite{kahana2022improving} & \multirow{3}*{Unlabeled} & 65.05 & 71.13 & 67.96 \\
        UPL~\cite{huang2022unsupervised}   & ~        & 74.83 & 78.04 & 76.40 \\
        LaFTer~\cite{mirza2023lafter}      & ~        & 79.49 & 82.91 & 81.16 \\
        \hline\noalign{\smallskip}
        FPL~\cite{menghini2023enhancing}   &          & 97.60 & 78.27 & 86.87 \\
        IFPL~\cite{menghini2023enhancing}  & Few-shot & 97.73 & 80.27 & 88.14 \\
        GRIP~\cite{menghini2023enhancing}  & +        & 97.83 & 80.87 & 88.54 \\
        \cellcolor{lightgray!30}PromptKD & Unlabeled & \cellcolor{lightgray!30}99.42 & \cellcolor{lightgray!30}82.62 & \cellcolor{lightgray!30}90.24 \\
        $\Delta$                           & ~         & \incre{+1.59} & \incre{+1.75} & \incre{+1.70} \\ 
        \hline
        \end{tabular}
    }
    \caption{Comparison with existing works using unlabeled data on the Flowers102 dataset. Our method performs better than previous methods.}
    \label{table:method_comparison}
    \vspace{-10pt}
\end{table}
\subsection{Ablation Study}
In this section, we perform ablation experiments on different components of the framework to verify its effectiveness. 
By default, the distillation experiments are conducted on the entire ImageNet. For experimental efficiency, we use 64 images per class by default, that is, a total of 64,000 images for 1,000 classes, as an unlabeled training set for distillation, unless we state otherwise. The accuracy of the base and new classes is evaluated on the test set. 

\begin{figure}[t]
    \centering
    \includegraphics[width=0.72\linewidth]{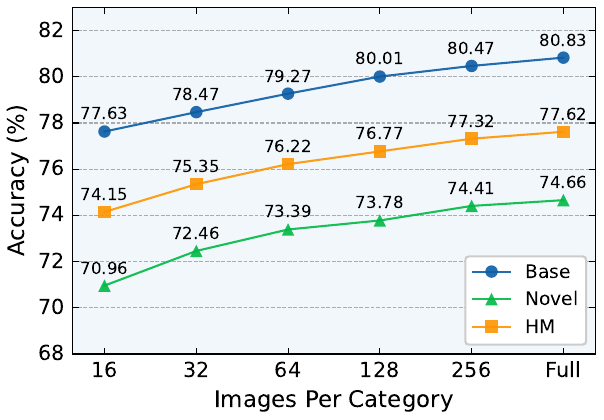}
    \vspace{-10pt}
    \caption{Improved ImageNet classification accuracy of the student model with increasing numbers of unlabeled images per category used for distillation.} 
    \label{figure:shots_in_training}
    \vspace{-8pt}
\end{figure}

\noindent\textbf{Number of Images Used for Training.}
In this section, our objective is to assess the influence of training data volume on distillation performance, as depicted in Fig.~\ref{figure:shots_in_training}. The figure illustrates that as the number of training images rises towards the complete training dataset, the accuracy consistently improves. It is noteworthy that with a further increase in the number of training images, the rate of performance improvement starts to plateau.


\begin{table}[ht]
    \centering
    \resizebox{0.65\linewidth}{!}
    {
        \begin{tabular}{ccccc}
        \hline\noalign{\smallskip}
        KD Form                & Loss & Base & Novel & HM    \\
        \hline\noalign{\smallskip}
        \multirow{2}*{Feature} & L1  & 73.09 & 65.98 & 69.35 \\
        ~                      & MSE & 71.89 & 66.17 & 68.91 \\
        \hline\noalign{\smallskip}
        Logit                  & KL  & \textbf{79.27} & \textbf{73.39} & \textbf{76.22}  \\
        \hline
        \end{tabular}
    }
    \caption{Comparison of different distillation forms. The logit-based form works best.
    }
    \label{table:kd_manner}
    \vspace{-10pt}
\end{table}

\noindent\textbf{Distillation Form.}
In Table~\ref{table:kd_manner}, we compare the performance of feature- and logit-based distillation.
In feature distillation, we align the features extracted by the teacher and student image encoders. Through careful hyperparameter tuning, we find that logit distillation yields significantly better results than the feature method. One possible reason is that the image feature space alignment is more difficult than the logit space alignment due to differences in the structure of the teacher and student models.



\noindent\textbf{Distillation Method.} In Table~\ref{table:ablation2}, we compare the performance of different distillation methods. We follow the same training settings of PromptKD to conduct the following experiments.
``Projector Only" represents that there is only the projector module in the student image encoder and no learnable prompts. ``Full Fine-tune" means that we fine-tune all parameters of the student CLIP model like CLIP-KD~\cite{yang2023clip}. ``w/o Shared Text Feature" means that we train the student model using its own text encoder along with learnable prompts to generate text features. The results indicate that the foundational designs of PromptKD, encompassing prompt-based distillation and shared class vectors from the teacher, play a crucial role in determining the ultimate performance.

\begin{table}[ht]
    \centering
    \resizebox{0.77\linewidth}{!}
    {
        \begin{tabular}{lccc}
        \hline\noalign{\smallskip}
        Method & Base  & Novel & HM   \\
        \hline\noalign{\smallskip}
        CLIP & 72.43 & 68.14 & 70.22 \\
        \hline\noalign{\smallskip}
        Projector Only & 78.48 & 72.79 & 75.53 \\
        Full Fine-tune
        & 75.90 & 70.95 & 73.34 \\
        w/o Shared Text Feature & 78.79 & 73.37 & 75.98 \\ 
        \hline\noalign{\smallskip}
        PromptKD & 79.27 & 73.39 & 76.22 \\
        \hline
        \end{tabular}
    }
    \caption{Ablation study of different distillation ways.}
    \label{table:ablation2}
    \vspace{-10pt}
\end{table}

\noindent\textbf{Teacher Pre-training Method.}
In Table~\ref{table:pretrain_manner}, we conduct experiments employing various methods for teacher pre-training, such as vanilla CLIP and MaPLe. The table illustrates that a higher accuracy attained by the teacher model through pre-training aligns with the improved distillation performance of the student model. Notably, any type of teacher model can enhance the student model with a non-trivial improvement.

\begin{table}[ht]
    \centering
    \resizebox{0.94\linewidth}{!}
    {
        \begin{tabular}{lcccc}
        \hline\noalign{\smallskip}
        Role~(Method) & Img Backbone & Base  & Novel & HM \\
        \hline\noalign{\smallskip}
        CLIP      & ViT-B/16   & 72.43 & 68.14 & 70.22 \\
        PromptSRC & ViT-B/16   & 77.60 & 70.73 & \cellcolor{lightgray!30}74.01 \\
        \hline\noalign{\smallskip}
        Teacher~(CLIP) & ViT-L/14 & 79.18 & 74.03 & 76.52 \\
        Student        & ViT-B/16 & 76.53 & 72.58 & 74.50 \\
        \hline\noalign{\smallskip}
        Teacher~(MaPLe) & ViT-L/14 & 82.79 & 76.88 & 79.73 \\
        Student         & ViT-B/16 & 78.43 & \textbf{73.61} & 75.95 \\  
        \hline\noalign{\smallskip}
        Teacher~(PromptSRC) & ViT-L/14 & 83.24 & 76.83 & 79.91 \\
        Student             & ViT-B/16 & \textbf{79.27} & 73.39 & \textbf{76.22} \\
        \hline
        \end{tabular}
    }
    \caption{Comparison of different pre-training methods. Teacher pre-training with PromptSRC brings the best student performance.
    }
    \label{table:pretrain_manner}
    \vspace{-10pt}
\end{table}

\begin{figure}[t]
    \centering
    \includegraphics[width=0.65\linewidth]{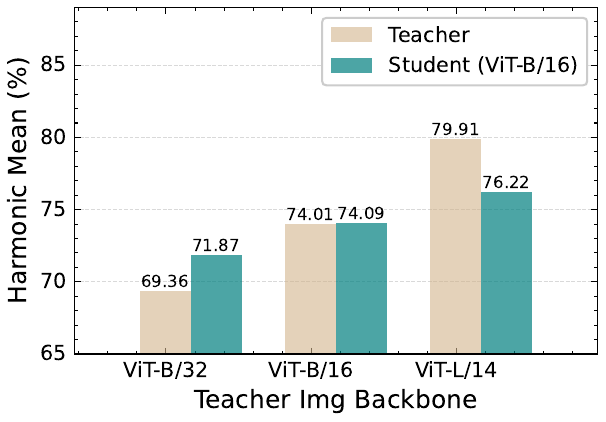}
    \vspace{-7pt}
    \caption{Comparison of distillation results for teachers with different capacities. Better teachers lead to better distillation performance.} 
    \label{figure:different_teacher}
    \vspace{-13pt}
\end{figure}

\noindent\textbf{Distillation with Different Pre-trained Teachers.}
In this part, we investigate the impact of using teacher models with different capacities on the performance of the student models, as shown in Fig.~\ref{figure:different_teacher}. We adopt the ViT-B/16 and ViT-B/32 CLIP models using the official PromptSRC code and employ them as pre-trained teacher models. 
The results indicate that stronger teacher models lead to better performance in distillation. 


\noindent\textbf{Inference Cost Analysis.} In Table~\ref{table:train_inference_cost}, we show the inference cost analysis and compare it with other prompt learning methods including CoOp, CoCoOp, and PromptSRC. The inference cost for all methods is calculated on a single A100 GPU on the SUN397 dataset. The results indicate that our method is more efficient than previous methods during inference, affirming its practicality in real-world applications.

\begin{table}[ht]
    \centering
    \resizebox{0.72\linewidth}{!}
    {
        \begin{tabular}{cccccc}
        \hline\noalign{\smallskip}
        Method & GFLOPs~(test) & FPS & HM \\
        \hline\noalign{\smallskip}
        CoOp      & 162.5  & 1344  & 71.66 \\
        CoCoOp    & 162.5  & 15.08 & 75.83 \\
        PromptSRC & 162.8  & 1380  & 79.97 \\
        \hline\noalign{\smallskip}
        PromptKD  & 42.5   & 1710  & 83.73 \\
        \hline
        \end{tabular}
    }
    \caption{Comparison of computation costs among existing methods on the SUN397 dataset. 
    Our PromptKD is more efficient than previous methods during testing.}
    \label{table:train_inference_cost}
    \vspace{-10pt}
\end{table}

\section{Conclusion}

In this paper, we introduce a two-stage unsupervised prompt distillation framework for Vision-Language Models, which aims to transfer the knowledge of a large CLIP teacher model to a lightweight CLIP student model through prompt imitation using unlabeled domain data.
Our method first pre-trains a large teacher model on domain few-shot labeled data and then performs student prompt distillation on extensive unlabeled domain data.
By leveraging CLIP's unique decoupled-modality property, we propose to reuse pre-stored teacher text features and incorporate them into the student image encoder for both distillation and inference purposes. Extensive experiments on 11 recognition datasets demonstrate the effectiveness of our method.

\noindent\textbf{Limitations and future work.} The effectiveness of the distillation method is intricately tied to the knowledge transferred through unlabeled domain samples. When the distillation data lacks representation from the target domain, the generalization capability of the distilled student model towards that specific domain may be biased or weakened. In the future, we plan to explore potential regularization methods to mitigate these issues. 


\noindent\textbf{Acknowledgement.} The authors would like to thank the anonymous reviewers for their critical comments and suggestions. The authors would also like to thank Shuheng Shen, Changhao Zhang, and Xing Fu for their discussions and help. This work was supported by the Young Scientists Fund of the National Natural Science Foundation of China~(No.62206134), the National Natural Science Fund of China~(No. 62361166670), the Fundamental Research Funds for the Central Universities 070-63233084 and the Tianjin Key Laboratory of Visual Computing and Intelligent Perception~(VCIP). Computation is supported by the Supercomputing Center of Nankai University~(NKSC).

{
    \small
    \bibliographystyle{ieeenat_fullname}
    \bibliography{main.bib}
}

\clearpage
\setcounter{page}{1}
\maketitlesupplementary

\section{Experimental Settings}
\noindent\textbf{Dataset.} We evaluate the performance of our method on 15 recognition datasets. For generalization from base-to-novel classes and cross-dataset evaluation, we evaluate the performance of our method on 11 diverse recognition datasets. Specifically, these datasets include ImageNet-1K~\cite{deng2009imagenet} and Caltech101~\cite{fei2004learning} for generic object classification; OxfordPets~\cite{parkhi2012cats}, StanfordCars~\cite{krause20133d}, Flowers102~\cite{nilsback2008automated}, Food101~\cite{bossard2014food}, and FGVCAircraft~\cite{maji2013fine} for fine-grained classification, SUN397~\cite{xiao2010sun} for scene recognition, UCF101~\cite{soomro2012ucf101} for action recognition, DTD~\cite{cimpoi2014describing} for texture classification, and EuroSAT~\cite{helber2019eurosat} for satellite imagery recognition. For domain generalization experiments, we use ImageNet-1K as the source dataset and its four variants as target datasets including ImageNet-V2~\cite{recht2019imagenet}, ImageNet-Sketch~\cite{wang2019learning}, ImageNet-A~\cite{hendrycks2021natural}, and ImageNet-R~\cite{hendrycks2021many}. 

\noindent\textbf{Training Details.}
For PromptKD, we follow the same settings as PromptSRC, setting the prompt depth to 9 and the vision and language prompt lengths to 4. We use the stochastic gradient descents~(SGD) as the optimizer. All student models are trained for 20 epochs with a batch size of 8 and a learning rate of 0.005. We follow the standard data augmentation scheme as in PromptSRC, i.e., random resized cropping and random flipping. The temperature hyperparameter $\tau$ in the current distillation method is default set to 1.  The text prompts of the first layer are initialized with the word embeddings of ``a photo of a \{classname\}”. We conduct all experiments on a single Nvidia A100 GPU. 

\noindent\textbf{Training Data Usage.}
In the initial stage of our method, we employ PromptSRC to pre-train our ViT-L/14 CLIP teacher model. During this stage, we utilize the same training data as PromptSRC for the training process. In the subsequent stage, we adopt the transductive zero-shot learning paradigm and employ the entire training dataset to train our student model. In Table~\ref{table:dataset_details}, we provide the details of the number of images used for training on the base-to-novel generalization setting.

\begin{table}[ht]
    \centering
    \resizebox{0.83\linewidth}{!}
    {
        \begin{tabular}{cccc}
        \hline\noalign{\smallskip}
        Dataset    & Train     & Test Base & Test Novel \\
        \hline\noalign{\smallskip}
        ImageNet      & 1,281,167 & 25,000 & 25,000 \\
        Caltech101    & 4,128     & 1,549  & 916    \\
        OxfordPets    & 2,944     & 1,881  & 1,788  \\
        StandfordCars & 6,509  & 4,002  & 4,039  \\
        Flowers102    & 4,093  & 1,053  & 1,410  \\
        Food101       & 50,500 & 15,300 & 15,000 \\
        FGVCAircraft  & 3,334  & 1,666  & 1,667  \\
        SUN397        & 15,880 & 9,950  & 9,900  \\
        DTD           & 2,820  & 864    & 828    \\
        EuroSAT       & 13,500 & 4,200  & 3,900  \\
        UCF101        & 7,639  & 1,934  & 1,849  \\
        \hline
        \end{tabular}
    }
    \caption{Number of images used for distillation and testing per dataset.}
    \label{table:dataset_details}
\end{table}


\section{Additional Experiments}

\noindent\textbf{Domain Generalization.}
In our PromptKD, the teacher model is first pre-trained using PromptSRC~\cite{khattak2023self} on the source dataset~(i.e., ImageNet). 
Then we train student models using unlabeled target datasets and then evaluate their performance after training.

In Table~\ref{table:domain_dataset}, we present the results of PromptKD and other state-of-the-art methods~(i.e., CoOp~\cite{zhou2022learning}, CoCoOp~\cite{zhou2022conditional}, MaPLe~\cite{khattak2023maple}, PromptSRC~\cite{khattak2023self}, TPT~\cite{shu2022test}, PromptAlign~\cite{samadh2023align}) on four different datasets. On the target dataset, our method shows a clear performance advantage compared to other methods.

\begin{table}[t]
    \centering
    \resizebox{1\linewidth}{!}
    {
        \begin{tabular}{cccccccc}
        \hline\noalign{\smallskip}
        ~    & ~         & \multicolumn{5}{c}{Target Dataset} \\
        ZSL  & ViT-B/16  & -V2 & -S & -A & -R & Avg. \\
        \hline\noalign{\smallskip}
        ~       & CLIP      & 60.83 & 46.15 & 47.77 & 73.96 & 57.18 \\
        ~       & CoOp      & 64.20 & 47.99 & 49.71 & 75.21 & 59.28 \\
        In-     & CoCoOp    & 64.07 & 48.75 & 50.63 & 76.18 & 59.91 \\
        ductive & MaPLe     & 64.07 & 49.15 & 50.90 & 76.98 & 60.27 \\
        ~       & PromptSRC & 64.35 & 49.55 & 50.90 & 77.80 & 60.65 \\
        \hline\noalign{\smallskip}
        ~       & TPT         & 63.45 & 47.94 & 54.77 & 77.06 & 60.81 \\
        ~       & CoOp+TPT    & 66.83 & 49.29 & 57.95 & 77.27 & 62.83 \\
        Trans-  & CoCoOp+TPT  & 64.85 & 48.47 & 58.47 & 78.65 & 62.61 \\
        ductive & PromptAlign & 65.29 & 50.23 & 59.37 & 79.33 & 63.55 \\
        ~       & \textbf{PromptKD} & \textbf{69.77} & \textbf{58.72} & \textbf{70.36}  & \textbf{87.01} & \textbf{71.47} \\
        ~       & $\Delta$ & \incre{+4.48} & \incre{+8.49} & \incre{+10.99} & \incre{+7.68} & \incre{+7.92} \\
        \hline
        \end{tabular}
    }
    \caption{Comparison of PromptKD with existing advanced approaches on domain generalization setting. Based on our pipeline, we perform unsupervised prompt distillation using the unlabeled domain data respectively (i.e., the transductive setting). The source model is training from ImageNet~\cite{deng2009imagenet}. ``ZSL'' denotes the setting type for Zero-Shot Learning. PromptKD achieves consistent improvement on all target datasets. }
    \label{table:domain_dataset}
\end{table}


\noindent\textbf{Teacher Accuracy.}
In Table~\ref{table:teacher_on_base2novel} and Table~\ref{table:cross_dataset_tea_acc}, we present the pre-trained ViT-L/14 based CLIP teacher model accuracy on the base-to-novel and cross dataset experiments.



\begin{table}[ht]
    \centering
    \resizebox{0.67\linewidth}{!}
    {
        \begin{tabular}{cccc}
        \hline\noalign{\smallskip}
        Dataset       & Base  & Novel & HM    \\
        \hline\noalign{\smallskip}
        ImageNet      & 83.24 & 76.83 & 79.91 \\
        Caltech101    & 98.71 & 98.03 & 98.37 \\
        OxfordPets    & 96.86 & 98.82 & 97.83 \\
        StandfordCars & 84.53 & 84.25 & 84.39 \\
        Flowers102    & 99.05 & 82.60 & 90.08 \\
        Food101       & 94.56 & 95.15 & 94.85 \\
        FGVCAircraft  & 54.44 & 43.07 & 48.09 \\
        SUN397        & 84.97 & 81.09 & 82.98 \\
        DTD           & 85.76 & 70.65 & 77.48 \\
        EuroSAT       & 94.79 & 83.15 & 88.59 \\
        UCF101        & 89.50 & 82.26 & 85.73 \\
        \hline
        \end{tabular}
    }
    \caption{Pre-trained ViT-L/14 CLIP teacher accuracy on base-to-novel generalization experiments.}
    \label{table:teacher_on_base2novel}
\end{table}


\begin{table}[t]
    \centering
    \resizebox{0.62\linewidth}{!}
    {
        \begin{tabular}{ccc}
        \hline\noalign{\smallskip}
        ViT-L/14 & Dataset      & Accuracy \\
        \hline\noalign{\smallskip}
        Source   & ImageNet     & 78.12 \\
        \hline\noalign{\smallskip}
        \multirow{10}*{Target} & Caltech101 & 95.61 \\
        ~       & OxfordPets    & 94.19 \\
        ~       & StandfordCars & 77.70 \\
        ~       & Flowers102    & 77.54 \\
        ~       & Food101       & 91.59 \\
        ~       & FGVCAircraft  & 31.29 \\
        ~       & SUN397        & 70.86 \\
        ~       & DTD           & 56.32 \\
        ~       & EuroSAT       & 47.55 \\
        ~       & UCF101        & 76.20 \\
        \hline\noalign{\smallskip}
        Avg.    &               & 71.89 \\
        \hline
        \end{tabular}
    }
    \caption{Pre-trained ViT-L/14 CLIP teacher accuracy on cross-dataset generalization experiments.}
    \label{table:cross_dataset_tea_acc}
\end{table}

\noindent\textbf{Layer of Projector.}
Table~\ref{table:projector} presents the distillation performance of different MLP layers used in the projector. The results show that two layers of MLP are effective enough to achieve feature alignment. More or fewer MLP layers will cause over-fitting or under-fitting problems in training.



\begin{table}[ht]
    \centering
    \resizebox{0.62\linewidth}{!}
    {
        \begin{tabular}{cccc}
        \hline\noalign{\smallskip}
        MLP Layer & Base            & Novel          & HM    \\
        \hline\noalign{\smallskip}
        1          & 78.97          & 72.90          & 75.81 \\
        \textbf{2} & \textbf{79.27} & \textbf{73.39} & \textbf{76.22} \\
        3          & 79.10          & 72.72          & 75.78 \\
        \hline
        \end{tabular}
    }
    \caption{Number of Projector layers. 2-layer MLP works best. 
    }
    \label{table:projector}
\end{table}

\begin{table}[ht]
    \centering
    \resizebox{0.82\linewidth}{!}
    {
        \begin{tabular}{ccccc}
        \hline\noalign{\smallskip} 
        Role     & Img Backbone & Base  & Novel & HM \\
        \hline\noalign{\smallskip}
        Teacher  & ViT-L/14     & 83.24 & 76.83 & 79.91 \\
        \hline\noalign{\smallskip}
        Baseline & \multirow{3}*{ViT-B/32} & 67.52 & 64.04 & 65.73 \\
        Student  & ~                       & 74.29 & 69.29 & 71.70 \\
        $\Delta$ & ~ & \incre{+6.77} & \incre{+5.25} & \incre{+5.97} \\
        \hline\noalign{\smallskip}
        Baseline & \multirow{3}*{ViT-B/16} & 72.43 & 68.14 & 70.22 \\
        Student  & ~                       & 80.83 & 74.66 & 77.62 \\
        $\Delta$ & ~ & \incre{+8.40} & \incre{+6.52} & \incre{+7.40} \\
        \hline
        \end{tabular}
    }
    \caption{Prompt distillation with different student CLIP models. $\Delta$ denotes the performance improvement compared to the baseline result. Student models of different capacities achieved consistent improvements.}
    \label{table:differrnt_stu}
\end{table}

\noindent\textbf{Distillation with Different Students.} 
To verify the effectiveness of PromtpKD on student models with different capacities, as shown in Table~\ref{table:differrnt_stu}, we further conduct experiments on the CLIP models with ViT-B/32 image encoder. The results show that the student models achieve consistent improvements through the PromptKD method.

\begin{figure}[ht]
    \centering
    \includegraphics[width=0.7\linewidth]{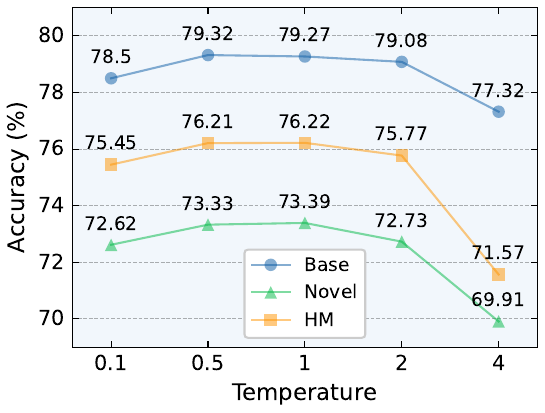}
    \vspace{-5pt}
    \caption{Choice of temperature hyperparameter. The best performance is achieved when $\tau$=1.} 
    \label{figure:temperature}
\end{figure}

\noindent\textbf{Temperature Hyperparameter.} 
The temperature parameter controls the softness of probability distribution~\cite{hinton2015distilling} and the learning difficulty of the distillation process~\cite{li2023curriculum}. In traditional distillation approaches, a common practice is to set the temperature parameter $\tau$ to 4 for most teacher-student pairs and datasets. In Fig.~\ref{figure:temperature}, we evaluate the impact of different temperature values on our proposed prompt distillation method. The results indicate that the traditional temperature setting of $\tau$=4 is not suitable for our current task. Increasing the temperature value leads to a rapid decrease in model performance. Interestingly, the best performance is achieved when $\tau$=1.

\noindent\textbf{Distillation with Longer Schedules.}
In PromptKD, for fair comparison, we adopt the same training schedule as PromptSRC, which is 20 epochs. In this part, we examine whether the student model can benefit from longer training schedules. As shown in Table~\ref{table:train_time}, we conduct experiments using 20, 40, and 60 training epochs respectively. The results show that the longer the training time, the higher the student performance.

\begin{table}[ht]
    \centering
    \resizebox{0.63\linewidth}{!}
    {
        \begin{tabular}{cccc}
        \hline\noalign{\smallskip}
        Train Epoch & Base  & Novel & HM    \\
        \hline\noalign{\smallskip}
        20          & 79.27 & 73.39 & 76.22 \\
        \hline\noalign{\smallskip}
        40          & 79.75 & 73.65 & 76.58 \\
        60          & \textbf{79.89} & \textbf{73.68} & \textbf{76.66} \\
        \hline
        \end{tabular}
    }
    \caption{Distillation with longer schedules. The longer the training time, the higher the student performance.}
    \label{table:train_time}
\end{table}

\section{Discussion}

\textbf{Experimental results of full fine-tune.}
In Table 5 of the main paper, we notice that the results of the full fine-tune method are lower than that of other distillation methods by a large margin~($>$2\%). There are two reasons for this. The first one is due to the limited size of the dataset we used in training. It is much smaller than the CC3M~\cite{sharma2018conceptual}, CC12M~\cite{changpinyo2021conceptual}, or LAION-400M~\cite{schuhmann2021laion} datasets commonly used to train CLIP. The second reason is that the training time is short. To align with other experimental settings, we only train the student model for 20 epochs. In total, the full fine-tuning method will improve if larger data sets are used and longer training schedules are adopted.

\noindent\textbf{Distillation with bad teachers.}
In Figure 5 of the main paper, when a weaker teacher~(ViT-B/32) is chosen compared to the student (ViT-B/16), the student trained using PromptKD demonstrates superior performance compared to the baseline method~(71.87\%$>$70.22\%). This situation differs from traditional distillation methods, where poor teachers often lead to a significant decline in student performance. The distinction arises due to the prompt learning method's focus on training only learnable prompts while keeping the original CLIP model weights frozen. The frozen CLIP model remains influential in the prediction process, where the trained prompts do not substantially bias the model inference.

\end{document}